\documentclass[10pt,twocolumn,letterpaper]{article}

\usepackage{iccv}
\usepackage{times}
\usepackage{epsfig}
\usepackage{graphicx}
\usepackage{amsmath}
\usepackage{amssymb}
\usepackage{algorithm}
\usepackage{algorithmic}
\usepackage{xcolor}
\usepackage{url}
\usepackage{footmisc}

\usepackage[breaklinks=true,bookmarks=false]{hyperref}

\iccvfinalcopy 

\def\iccvPaperID{2290} 

\def\Max{\textsf{MAX}}
\ificcvfinal\fi
\input{Definitions}
\graphicspath{{figures/}}

\begin{document}

\title{PointCloud Saliency Maps}

\author{Tianhang Zheng \qquad
Changyou Chen \qquad Junsong Yuan\\
State University of New York at Buffalo\\
{\tt\small \{tzheng4, changyou, jsyuan\}@buffalo.edu}
\and
Bo Li \qquad Kui Ren\\
Zhejiang University\\
{\tt\small \{boli, kuiren\}@zju.edu.cn}
}

\maketitle
\ificcvfinal\thispagestyle{empty}\fi

\begin{abstract}
3D point-cloud recognition with PointNet and its variants has received remarkable progress. A missing ingredient, however, is the ability to automatically evaluate point-wise importance w.r.t.\! classification performance, which is usually reflected by a saliency map. A saliency map is an important tool as it allows one to perform further processes on point-cloud data. In this paper, we propose a novel way of characterizing critical points and segments to build point-cloud saliency maps. Our method assigns each point a score reflecting its contribution to the model-recognition loss. The saliency map explicitly explains which points are the key for model recognition. Furthermore, aggregations of highly-scored points indicate important segments/subsets in a point-cloud. Our motivation for constructing a saliency map is by point dropping, which is a non-differentiable operator. To overcome this issue, we approximate point-dropping with a differentiable procedure of shifting points towards the cloud centroid. Consequently, each saliency score can be efficiently measured by the corresponding gradient of the loss w.r.t the point under the spherical coordinates.  
	 Extensive evaluations on several state-of-the-art point-cloud recognition models, including PointNet, PointNet++ and DGCNN, demonstrate the veracity and generality of our proposed saliency map. Code for experiments is released on \url{https://github.com/tianzheng4/PointCloud-Saliency-Maps}.
\end{abstract}

\section{Introduction}

Point clouds, which comprise raw outputs of many 3D data acquisition devices such as radars and sonars, are an important 3D data representation for computer-vision applications \cite{linsen2001point,vosselman2004recognising,rusu2008towards,rusu2009fast}. Real applications such as object classification and segmentation usually require high-level processing of 3D point clouds \cite{thrun2006probabilistic,hadsell2009learning,biswas2012depth,kehoe2013cloud}. Recent research has proposed to employ Deep Neural Network (DNN) for high-accuracy and high-level processing of point clouds, achieving remarkable success. Representative DNN models for point-cloud data classification include PointNet \cite{qi2017pointnet}, PointNet++ \cite{qi2017pointnet++} and DGCNN \cite{wang2018dynamic}, which successfully handled the irregularity of point clouds and achieved high classification accuracy. 
\begin{figure}
	\centering
	\includegraphics[width=0.95\columnwidth]{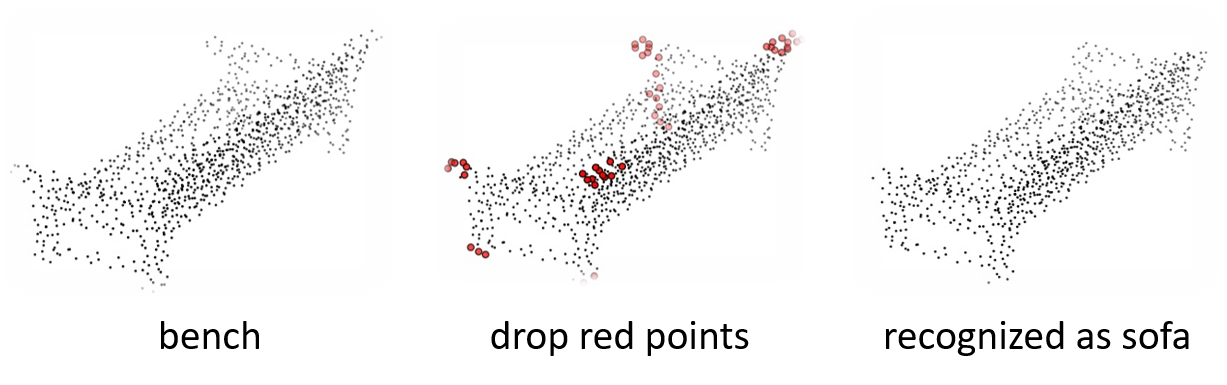}
	
	\caption{Drop the $5\%$ most critical points identified by our saliency map from a bench point cloud can easily change the prediction outcome ({\em even can trick human vision!}).}
	\label{fig:first_example}
	\vspace{-0.3cm}
\end{figure}
Beyond that, a notable characteristic of PointNet and its variants is their robustness to furthest/random point dropping. \cite{qi2017pointnet} owes the robustness to the max pooling layer in PointNet, which only concentrates on a critical subset of a point cloud. In other words, the recognition result is mainly determined by those critical points such that dropping some other non-critical points does not change the prediction. We refer to the corresponding theory given in \cite{qi2017pointnet} as critical-subset theory.
{\em Despite identifying such an important subset, we observed that the critical-subset theory is too ambiguous, as it does not specify the importance of each point and subset.} In this paper, we propose a simple method to construct a general saliency map for point-level and subset-level saliency assessment. Note in \cite{simonyan2013deep,springenberg2014striving,papernot2016limitations}, saliency map is constructed for images to characterize the contribution of each pixel value to the recognition result. We extend this concept to point cloud, aiming at studying importance of each {\em single point}.
Specifically, our method assigns a saliency score for each point, reflecting the contribution of a point to the corresponding model-prediction loss. {\em A saliency map is important to better understand point-cloud data, in that: On the one hand, if one drops points with the highest saliency scores, model performance would decrease significantly, endowing the potential to build an adversarial attack model. On the other hand, if only points with the lowest scores are dropped, model performance would not change a lot. Somewhat surprisingly, we find dropping points with negative scores even leads to better recognition performance.}

Despite simplicity in concept, how to construct such a point-level saliency map is nontrivial. One possible solution is to {\em drop all possible combinations of points and compute the loss changes after dropping those combinations, {\it i.e.}, loss difference caused by those combinations.} However, this simple brute-force method is impractical because the computational complexity scales exponentially w.r.t.\! the number of points in a point cloud. Instead, we propose an efficient and effective method to approximate saliency maps with a single backward step through DNN models. The basic idea is to approximate point dropping with a continuous point-shifting procedure, {\it i.e.,} moving points towards the point-cloud center. This is intuitively valid because the point cloud center is supposed to be uninformative for classification. In this way, prediction-loss changes can be approximated by the gradient of the loss w.r.t.\! the point under a spherical coordinate system. Thus, every point in a point cloud is associated with a score proportional to the gradient of loss w.r.t.\! the point. {\em We further propose an iterative point-dropping algorithm for verification of our saliency map. As stated above, if our saliency map is effective, dropping points with the highest (positive)/lowest (negative) saliency scores will degrade/improve model performance.} Surprisingly, some point clouds manipulated by our point-dropping algorithm even concur with human intuition well as shown in Fig.~\ref{fig:first_example}, indicating our saliency map can recognize salient points and segments like human does.

We compared our saliency-map-driven point-dropping algorithms with the random point-dropping baseline and the best critical-subset-based strategy on several state-of-the-art point-cloud DNN models, including PointNet, PointNet++, and DGCNN. We show that our method can always outperform those schemes in terms of improving or degrading model performance with limited points dropped. As an example, we show that dropping $200/1024$ points with the highest saliency scores from each point cloud by our algorithm can reduce the classification accuracy of PointNet on 3D-MNIST/ModelNet40 to $49.2\%$/$44.3\%$, while the random-dropping scheme only reduce the accuracy to $94.8\%$/$87.7\%$, close to original accuracies. Besides, the best critical-subset-based strategy ({\em only applicable to PointNet}) only reduces the accuracies to $80.0\%$/$58.1\%$. {\em All those experiments verified that our saliency map is a more accurate way to characterize point-level and even subset-level saliency than the critical-subset theory.}

\section{Preliminaries}
\subsection{Definition and Notations}\label{sec:pointcloud}
\paragraph{Point Cloud} 
A point cloud is represented as $(\Xb \triangleq \{\xb_i\}_{i = 1...N},  y)$, where $\xb_i \in \mathbb{R}^3$ is a 3D point and $N$ is the number of points in the point cloud; $y \in \{1, 2,...,k\}$ is the ground-truth label, where $k$ is the number of classes. We denote the output of a point-cloud classification network as $\Fb_\thetab(\cdot) \triangleq \{F_{\thetab, j}| j = 1...k\}$, whose input is a point cloud $\Xb$ and output is a probability vector $\Fb_\thetab(\Xb)$. The classification loss of the network is denoted as $\mathcal{L}(\Xb, y; \thetab)$, which is usually defined as the cross-entropy between $\Fb_\thetab(\Xb)$ and $y$. 

\paragraph{Point Contribution}
{\em We define the contribution of a point/points in a point cloud as the difference between the prediction-losses of two point clouds including or excluding the point/points, respectively.} 
Formally, given a point $\xb_i$ in $\Xb$, the contribution of $\xb_i$ is defined as $\mathcal{L}(\Xb' , y; \thetab) - \mathcal{L}(\Xb, y; \thetab)$, where $\Xb' \triangleq \{\xb_j:j = 1...N, j \neq i\}$. If this value is positive (or large), we consider the contribution of $\xb_i$ to model prediction as positive (or large). Because in this case, if $\xb_i$ is added back to $\Xb'$, the loss will be reduced, leading to more accurate classification. Otherwise, we consider the contribution of $\xb_i$ to be negative (or small).
\paragraph{Image and Point-Cloud Saliency Map}
Existing works on model interpretation and vulnerability have constructed saliency maps for images to identify which pixels are critical to model-recognition and how those pixel values can influence the recognition performance \cite{simonyan2013deep,springenberg2014striving,papernot2016limitations}.
We first propose a similar saliency map for point cloud here. {\em Point-cloud saliency map assigns each point $\xb_i$ a saliency score, {\it i.e.,} $s_i$, to reflect the contribution of $\xb_i$.} Formally, the map can be denoted as a function $S_\thetab(\cdot)$ with input $\Xb$ and outputting a vector of length $N$, {\it i.e.}, $\{s_i| i = 1...N\}$. We expect higher (positive) $s_i$ to indicate more (positive) contribution of $\xb_i$. We can use point-dropping to verify the veracity of our saliency map.
\paragraph{Point Dropping}\label{subsec:pointdrop}
Point dropping is a method to evaluate the veracity of our proposed saliency map. 
If our saliency map is accurate, then dropping points with the highest(positive)/lowest(negative) saliency scores will degrade/improve recognition performance. Ideally, high (positive) saliency scores indicate significant positive contributions to the recognition result. Thus after dropping points with the highest scores, we are expected to have $\arg_j\max F_{\thetab, j}(\Xb') \neq y$, where $\Xb'$ is the remaining point cloud. On the contrary, especially when the dropped points have negative saliency scores, which means they contribute negatively to the prediction, we should have $\arg_j\max F_{\thetab, j}(\Xb') = y$. 

\subsection{3D Point-Cloud Recognition Models}\label{sec:3dmodel}
There are three mainstream approaches for 3D object recognition: volume-based \cite{wu20153d, maturana2015voxnet}, multi-view-based \cite{su2015multi, wang2017dominant, yu2018multi, kanezaki2018rotationnet}, and point-cloud-based \cite{qi2017pointnet, qi2017pointnet++, wang2018dynamic} approaches, which rely on voxel, multi-view-image, and point-cloud representations of 3D objects, respectively. In this work, we focus on point-cloud-based models.
\paragraph{PointNet and PointNet++}
PointNet \cite{qi2017pointnet} applies {\em a composition of single variable-functions, a max pooling layer, and a function of the max pooled features}, which is invariant to point orders, to approximate the functions for point-cloud classification and segmentation. Formally, the composition can be denoted as $\gamma \circ \underset{\xb_i \in \Xb}{\Max}\{\hb(\xb_i)\}$, with $\hb(\cdot)$ a single-variable function, $\Max$ the max-pooling layer, and $\gamma$ a function of the max pooled features ({\it i.e.,} ${\Max}\{\hb(\xb_i)\}$).
PointNet plays a significant role in the recent development of point-cloud high-level processing, serving as a baseline for many following point-cloud DNN models.
PointNet++ \cite{qi2017pointnet++} is one extension, which applies PointNet recursively on a nested partitioning of the input point set, to capture hierarchical structures induced by the metric space where points live in. Compared to PointNet, PointNet++ is able to learn hierarchical features w.r.t.\! the Euclidean distance metric, and thus typically achieves better performance.
\paragraph{Dynamic Graph Convolutional Neural Network (DGCNN)}
DGCNN \cite{wang2018dynamic} integrates a novel operation into PointNet, namely EdgeConv, to capture local geometric structures while maintaining network invariance to point-permutation. Specifically, the operation EdgeConv generates features that can describe the neighboring relationships by constructing a local neighborhood graph and applying convolutional-like operations on edges connecting neighboring pairs of points. EdgeConv helps DGCNN achieve further performance improvement, usually surpassing PointNet and PointNet++.
\paragraph{Critical-Subset Theory}
For any point cloud $\Xb$, \cite{qi2017pointnet} proves that there exists a subset $\Cb \subseteq \Xb$, namely critical subset, which determines all the max pooled features $\ub$, and thus the output of PointNet. We briefly explain this theory in the following: a PointNet network can be expressed as $\Fb(\Xb) \triangleq \gamma \circ \ub(\Xb)$, where $\gamma$ is a continuous function, and $\ub(\Xb)$ represents the max pooled features. Apparently, $\Fb(\Xb)$ is determined by $\ub(\Xb)$. $\ub(\Xb)$ is computed by $\ub(\Xb) = \underset{\xb_i \in \Xb}{\Max}\{\hb(\xb_i)\}$, where $\Max$ (i.e., a special maxpooling layer) is an operator that takes $N$ vectors as input and returns a new vector of the element-wise maximums. For the $jth$ dimension of $\ub$, there exists one $\xb_i \in \Xb$ such that $\ub_j = \hb_j(\xb_i)$, where $\hb_j$ is the $j-th$ dimension of $\hb$. Aggregate all those $\xb_i$ into a subset $\Cb \subseteq \Xb$ such that $\Cb$ will determine $\ub$ and thus $\gamma \circ \ub$. \cite{qi2017pointnet} named $\Cb$ as critical subset.
As we can see, this theory is applicable to network structures similar to $\mathbf{\gamma} \circ \underset{\xb_i \in \Xb}{\Max}\{\hb(\xb_i)\}$, where a max-pooled feature is simply determined by one point, but not to networks with more complicated structures.
Visually, $\Cb$ usually distributes evenly along the skeleton of $\Xb$. In this sense, for PointNet, the critical subset seems to include all the points critical to the recognition result. {\em We refer the readers who are interested in more details to the appendix in \cite{qi2017pointnet}.}
Although the critical-subset theory helps to identify a salient point subset, we found that the theory does not specify point-level saliency yet, and it is also not an accurate and exhaustive way to characterize subset-level saliency. 

\section{Point-Cloud Saliency Map}
In this section, we derive our proposed saliency map following the definitions in Section \ref{sec:pointcloud}. Instead of dropping every point/subset and calculating the loss change (difference), we approximate point dropping by the procedure of shifting points to the spherical core (center) of a point cloud. Through this way, the nondifferentiable loss change caused by point-dropping can be approximated by differentiable loss change under a point-shifting operation, based on which a saliency map is constructed.

\subsection{From Point Dropping to Point Shifting}\label{subsec:core_theory}
\begin{figure}
 	\centering
 	\includegraphics[width=0.8\columnwidth]{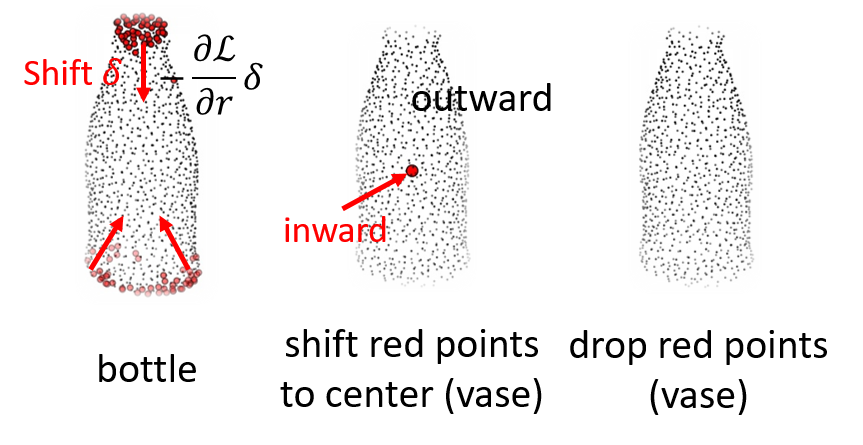}
 	\vspace{-0.2cm}
 	\caption{Approximate point dropping with point shifting toward the point-cloud center.}
 	\label{fig:core_theory}
 	\vspace{-0.3cm}
 \end{figure}

Our idea is illustrated in Fig.~\ref{fig:core_theory}. The intuition is that all the external (outward) points of a point cloud are supposed to determine the recognition result, because those points encode shape information of objects, while the points near the point center (inward) \footnote{Median value of x, y, z coordinates} almost have no effect on the recognition performance. {\em More concretely, Outward” corresponds to all original external points not shifted (to the center).}
{\em Consequently, dropping a point has similar effects to shifting the point towards the center in terms of eliminating the effect of the point on the classification result.} A more precise explanation for this intuition in theory is that the central points for all the point clouds are at the same position after coordinate translation so that their contribution to recognition can be neglected. Formally, we divide a point cloud into two parts $\{\{\xb'_i\}, \{\cbb_i\}\}$, where $\{\cbb_i\}$ represents the point subset at the centroid, and $\{\xb'_i\}$ represents the remaining points on the surface. For a natural point cloud, $\{\cbb_i\}$ is usually an empty set. The max-pooling layer in PointNet can be rewritten as ${\Max}\{\hb(\xb_i)\} = \max\{{\Max}\{\hb(\xb'_i)\}, {\Max}\{\hb(\cbb_i)\}\}$, where $\max(\ab, \bb)$ returns the element-wise maximum of $\ab$ and $\bb$. Since $\{\cbb_i\}$ is the same for all the point clouds after coordinate transformation, determinant max-pooled features should mainly come from ${\Max}\{\hb(\xb'_i)\}$. 

To verify our hypothesis, we conduct a proof-of-concept experiment: thousands of pairs of point clouds are generated by dropping $100/1024$ points and shifting those $100/1024$ points to the point cloud center respectively. Here we totally used three schemes to select those $100/1024$ points, including furthest point-dropping, random point-dropping, and point-dropping based on our saliency map. We use PointNet for classification of both of the point clouds in every pair. For all those selection schemes, the classification results achieve more than $95\%$ pairwise consistency\footnote{For more than $95\%$ pairs, the classification results of the two point clouds in each pair are the same (may be correct or wrong)}, indicating applicability of our approach.


\begin{figure*}
	\centering
	\includegraphics[width=0.48\columnwidth]{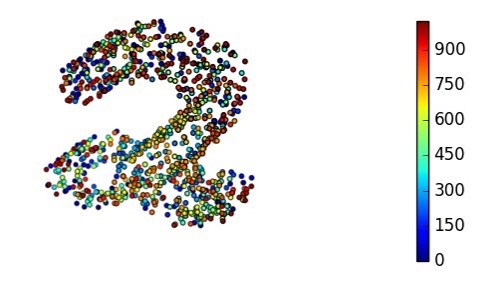}
	\includegraphics[width=0.48\columnwidth]{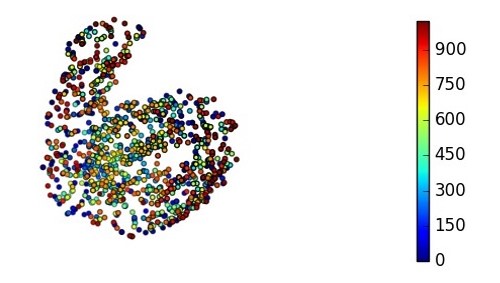}	
	\includegraphics[width=0.48\columnwidth]{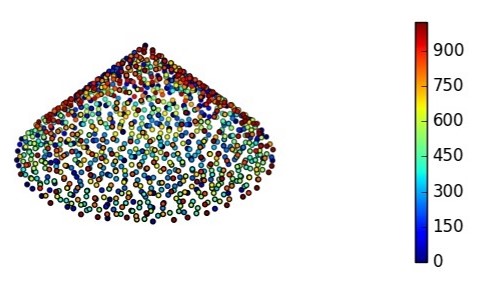}
	\includegraphics[width=0.48\columnwidth]{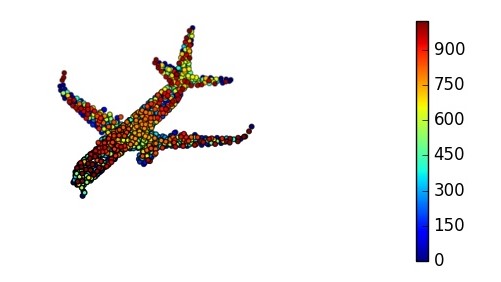}
	
	\includegraphics[width=0.48\columnwidth]{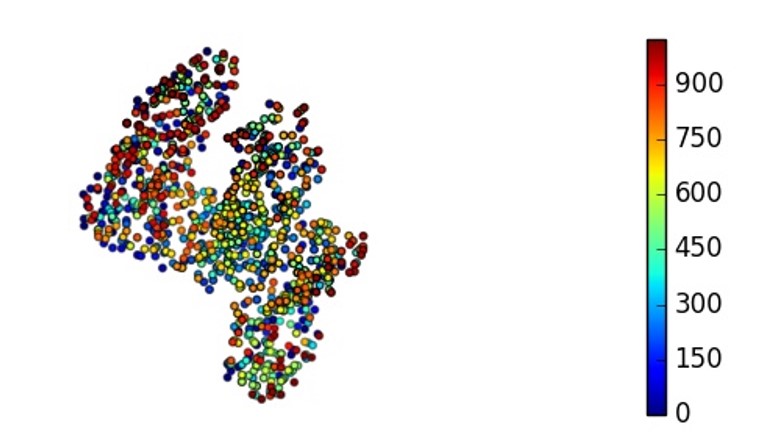}
	\includegraphics[width=0.48\columnwidth]{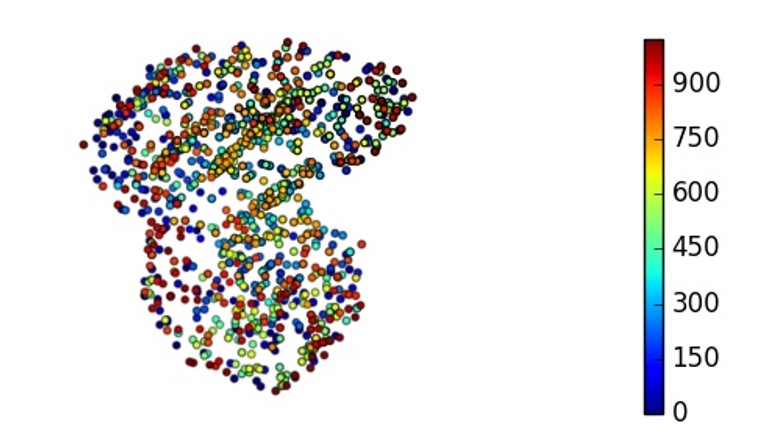}	
	\includegraphics[width=0.48\columnwidth]{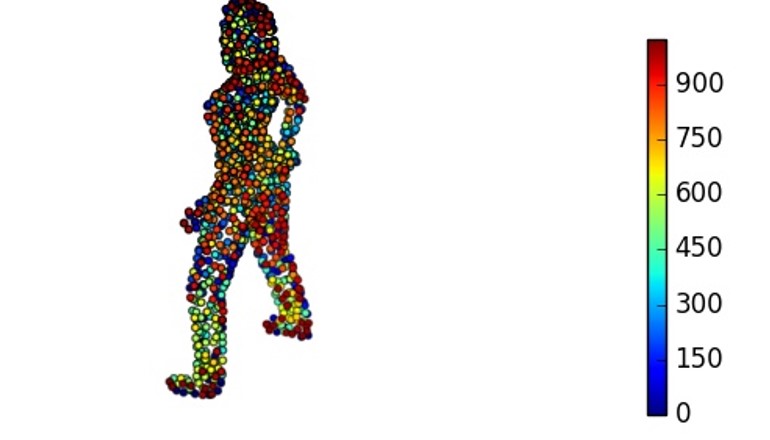}
	\includegraphics[width=0.48\columnwidth]{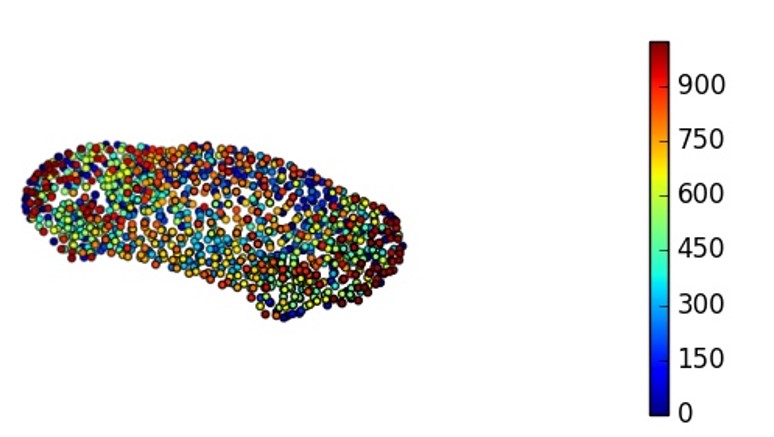}
	\caption{Visualize several saliency maps of digits and objectives (one-step): coloring points by their score-rankings.}
	\label{fig:saliency_map}
\end{figure*}
\subsection{Gradient-based Saliency Map}\label{subsec:saliency_map}
{\em Based on the intuition in \ref{subsec:core_theory}, we approximate the contribution of a point by the gradient of loss, i.e., the difference between the prediction-losses of two point clouds including or excluding the point, under the point-shifting operation.} Note that measuring gradients in the original coordinate system is problematic because points are not view (angle) invariant. In order to overcome this issue, we consider point shifting in the Spherical Coordinate System, where a point is represented as $(r, \psi, \phi)$ with $r$ distance of a point to the spherical core, $\psi$ and $\phi$ the two angles of a point relative to the spherical core. Under this spherical coordinate system, as shown in Fig. \ref{fig:core_theory}, shifting a point towards the center by $\delta$ will increase the loss $\mathcal{L}$ by $-\frac{\partial \mathcal{L}}{\partial r}\delta$. Then based on the equivalence we established in section \ref{subsec:core_theory}, we measure the contribution of a point by a real-valued score -- negative gradient of the loss $\mathcal{L}$ w.r.t. $r$, {\it i.e.,} $-\frac{\partial \mathcal{L}}{\partial r}$. 
To calculate $\frac{\partial \mathcal{L}}{\partial r}$ for certain point cloud, we use the medians of the axis values of all the points in the point cloud as the spherical core, denoted as $\xb_c$, to build the spherical coordinate system for outlier-robustness \cite{bohm2008outlier}. Formally, $\xb_c$ can be expressed as
\begin{align}\label{eq:core}
\xb_{cj} = median(\{\xb_{ij}|~ \xb_i \in \Xb \}) ~~ (j = 1, 2, 3),
\end{align}
where $(\xb_{i1}, \xb_{i2}, \xb_{i3})$ represent the axis values of point $\xb_i$ corresponding the orthogonal coordinates $(x, y, z)$.
Consequently, $\frac{\partial \mathcal{L}}{\partial r}$ can be computed by the gradients under the original orthogonal coordinates as: 
\begin{align}\label{eq:rgrad}
\frac{\partial \mathcal{L}}{\partial r_i} = \sum_{j=1}^{3} \frac{\partial \mathcal{L}}{\partial \xb_{ij}}\frac{\xb_{ij} - \xb_{cj}}{r_i},
\end{align}
where $r_i = \sqrt{\sum_{j=1}^{3}(\xb_{ij} - \xb_{cj})^2}$.
In practice, we apply a change-of-variable by $\rho_i = r_i^{-\alpha}$ ($\alpha > 0$) to allow more flexibility in saliency-map construction, where $\alpha$ is used to rescale the point clouds. The gradient of $\mathcal{L}$ w.r.t. $\rho_i$ can be calculated by
\vspace{-0.1cm}
\begin{align}\label{eq:rhograd}
\frac{\partial \mathcal{L}}{\partial \rho_i} = -\frac{1}{\alpha} \frac{\partial \mathcal{L}}{\partial r_i} r_i^{1+\alpha}.
\end{align}
Define $\delta_\rho$/$\delta_r$ as a differential step size along $\rho$/$r$. Since $\delta_\rho = -\alpha r^{-(\alpha + 1)}\delta_r$, shifting a point $(r, \psi, \phi)$ by $-\delta_r$ ({\it i.e., }$\delta_r$ towards the center $r=0$) is equivalent to shifting the point by $\delta_\rho$ if ignoring the positive factor $\alpha r^{-(\alpha + 1)}$. Therefore, {\em under the framework of $(\rho, \psi, \phi)$, we approximate the loss change by $\frac{\partial \mathcal{L}}{\partial \rho}\delta_\rho$, which is proportional to $\frac{\partial \mathcal{L}}{\partial \rho}$}. Thus in the rescaled coordinates, we measure the contribution of a point $\xb_i$ by $\frac{\partial \mathcal{L}}{\partial \rho_i}$, {\it i.e.,} $ -\frac{1}{\alpha} \frac{\partial \mathcal{L}}{\partial r_i} r_i^{1+\alpha}$. Since $\frac{1}{\alpha}$ is a constant, we simply employ
\vspace{-0.3cm}
\begin{align}\label{eq:score}
s_i = -\frac{\partial \mathcal{L}}{\partial r_i} r_i^{1+\alpha}
\end{align}
as the saliency score of $\xb_i$ in our saliency map. Note the additional parameter $\alpha$ gives us extra flexibility for saliency-map construction, and optimal choice of $\alpha$ would be problem specific. {\em In the following experiments, we simply set $\alpha$ to 1, which already achieves remarkable performance.}
For better understanding of our saliency maps, several maps are visualized in Fig. \ref{fig:saliency_map}. We colorcode those points by the ranks of their saliency scores, i.e., larger number indicated higher saliency scores.

\section{Point Dropping Algorithms}
As stated in Section \ref{subsec:pointdrop}, point dropping can be used to verify the veracity of our saliency map. Therefore, we propose two point dropping algorithms in Section \ref{subsec:salient_alg}. For comparison with the critical-subset theory, we also tried several critical-subset based point dropping strategies, and present the most effective one in Section \ref{subsec:subset_alg}. For simplicity, we refer to dropping points with the highest scores as  {\em high-drop}, dropping points with the lowest scores as {\em low-drop}, and the most effective critical-subset based strategy as {\em critical} in the followings.
Except for verification, point dropping is also helpful for understanding subset-level (segment-level) saliency. For instance, after {\em high-drop}, the remaining fragmented point cloud will be recognized as another object, which means the dropped points belong to the most important segments in the object for recognition. Surprisingly, the points dropped by our saliency-map based {\em high-drop} algorithms are always clustered as illustrated in Fig.~\ref{fig:high}, and the clusters are indeed the critical segments for object recognition even in human eyes.


\subsection{Saliency-Map based Point Dropping}\label{subsec:salient_alg}
Based on the illustrations in Section \ref{subsec:saliency_map}, saliency maps are readily constructed by calculating gradients following \eqref{eq:score}, which guide our point-dropping processes (algorithms). 
Algorithm \ref{alg:iterdrop} describes our iterative algorithm for point dropping. 
{\em Note calculating saliency scores at once might be suboptimal because point dependencies have been ignored}. 
To alleviate this issue, we propose to drop points iteratively such that point dependencies in the remaining point set will be considered when calculating saliency scores for the next iteration. Specifically, in each iteration, a new saliency map is constructed for the remaining points, and among them $n/T$ points are dropped based on the current saliency map. {\em In section \ref{sec:empirical}, we set $n/T = 5$ for dropping points with the highest saliency scores and show that this setting is good enough in terms of improving the performance and understanding subset-level saliency with reasonable computational cost.}
%
%
\begin{algorithm}
	\caption{Iteratively drop points based on dynamic saliency maps}
	\label{alg:iterdrop}
	\begin{algorithmic}
		\REQUIRE Loss function $\mathcal{L}(\Xb, y; \thetab)$; point cloud input $\Xb$, label $y$, and model weights $\thetab$; hyper-parameter $\alpha$; total number of points to drop $n$; number of iterations $T$.
		\FOR {$t$ = $0$ to $T$}
		\STATE Compute the gradient $\gb^t_i = \nabla_{\xb^t_i}\mathcal{L}(\Xb^t, y; \thetab)$
		\STATE Compute the center by $\xb^t_c \triangleq (x^t_{c1}, x^t_{c2}, x^t_{c3}) = median(\xb^t_{i1}, \xb^t_{i2}, \xb^t_{i3})$
		\STATE Compute $r_i\frac{\partial \mathcal{L}}{\partial r_i} = (\xb^t_i - \xb^t_c) \cdot \gb^t_i$ (inner product)
		\STATE Construct the saliency map by $ s_i = -r_i^{\alpha}r_i\frac{\partial \mathcal{L}}{\partial r_i}$
		\IF{high-drop}
		\STATE Drop the points with $n/T$ lowest $s_i$ from $\Xb^t$
		\ELSIF{low-drop}
		\STATE Drop the points with $n/T$ highest $s_i$ from $\Xb^t$
		\ENDIF
		\ENDFOR
		\STATE Output $\Xb^T$
	\end{algorithmic}
\end{algorithm}

\subsection{Critical-Subset based Point Dropping}\label{subsec:subset_alg}
To compare our saliency map with the critical-subset theory, we also propose several point-dropping strategies based on the critical-subset theory, e.g., randomly dropping points from the critical-subset one-time/iteratively and dropping the points that contribute to the most number of max-pooled features one-time/iteratively. Among all those critical-subset based schemes, dropping the points with contribution to the most number of max-pooled features (at least two features) iteratively provides the best performance. The strategy is illustrated in Algorithm \ref{alg:itercritical}. However, we found that even this scheme still performs worse than our saliency-map based point-dropping algorithm, which indicates that our saliency map is a more accurate measure on the point-level and subset-level saliency. 

\begin{algorithm}
	\caption{Iteratively drop points based on dynamic critical subset}
	\label{alg:itercritical}
	\begin{algorithmic}
		\REQUIRE PointNet network $f=\gamma \circ \underset{\xb_i \in \Xb}\Max\{h(\xb_i)\}$; point cloud input $\Xb$, label $y$, and model weights $\thetab$; hyper-parameter $\alpha$; total number of points to drop $n$; number of iterations $T$.
		\FOR {$t$ = $0$ to $T$}
		\STATE Compute the indexes of points in the critical-subset ({\em index list}) is by $\arg \Max \{h(\xb_i)\}$
		\STATE Count $c_i \triangleq ~\mbox{the frequency of}~ i ~\mbox{in the list}$ ({\em i.e.,} $\xb_i$ determines $c_i$ max-pooled features)
		\STATE Drop $n/T$ points with the largest $c_i$ from $\Xb^t$
		\ENDFOR
		\STATE Output $\Xb^T$
	\end{algorithmic}
\end{algorithm}
\begin{figure*}[t]	
	\centering
	\includegraphics[width=0.48\columnwidth]{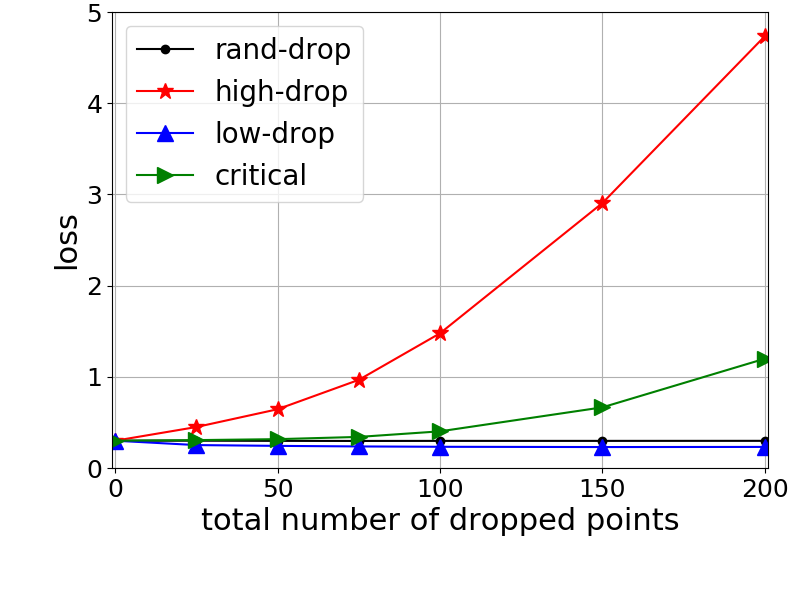}
	\includegraphics[width=0.48\columnwidth]{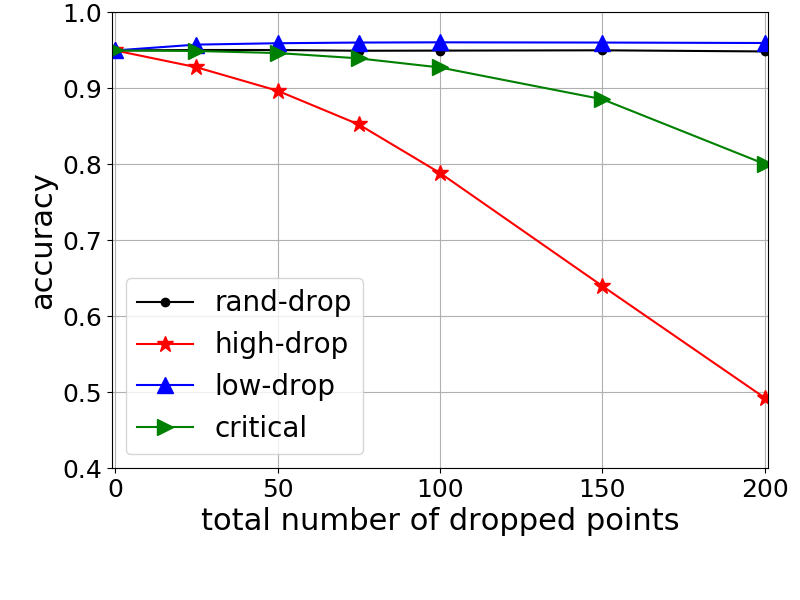}
	\includegraphics[width=0.48\columnwidth]{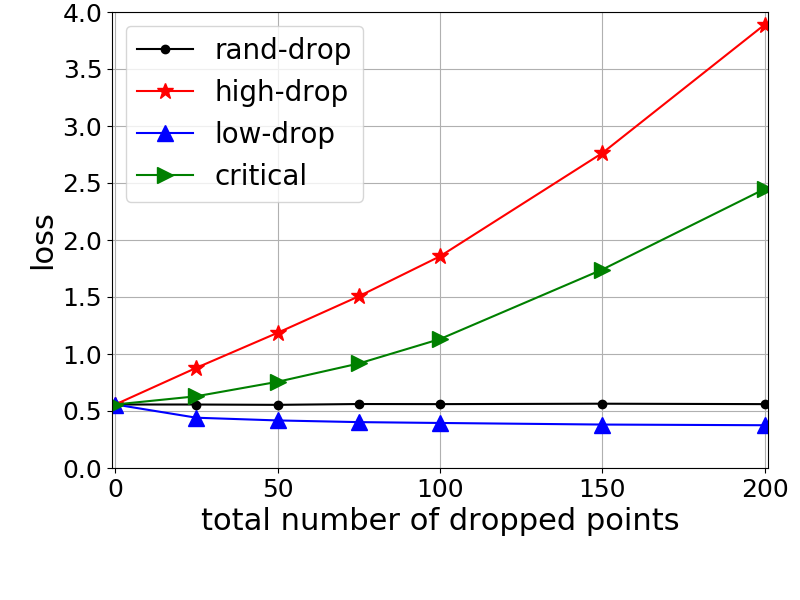}
	\includegraphics[width=0.48\columnwidth]{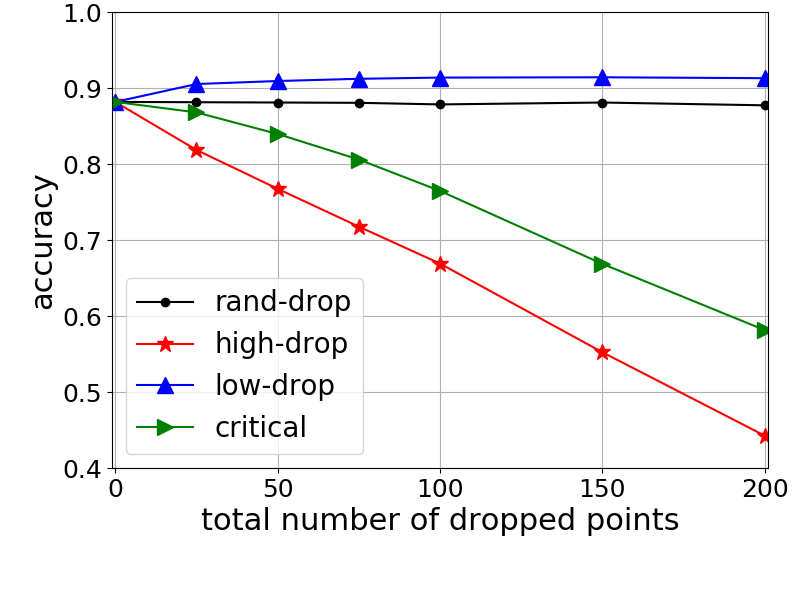}
	\vspace{-0.2cm}
	\caption{PointNet on 3D-MNIST and ModelNet40 from left to right: averaged loss (3D-MNIST), overall accuracy (3D-MNIST), averaged loss (ModelNet40), overall accuracy (ModelNet40).}
	\label{fig:pointnet}
	\vspace{-0.2cm}
\end{figure*}
\begin{figure*}[t]	
	\centering
	\includegraphics[width=0.48\columnwidth]{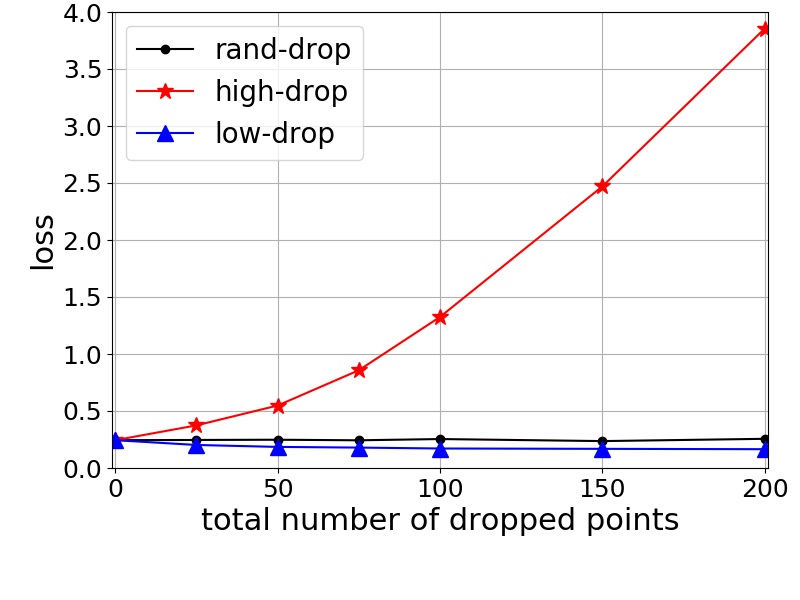}
	\includegraphics[width=0.48\columnwidth]{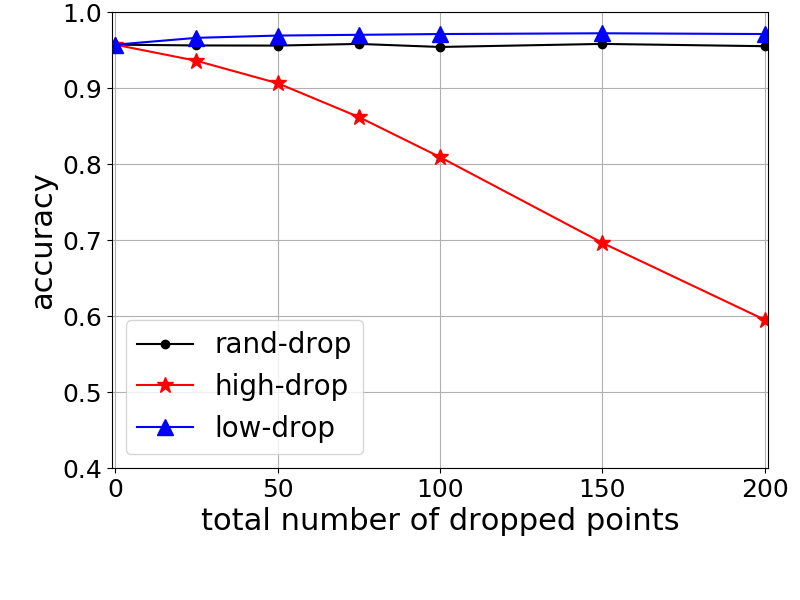}
	\includegraphics[width=0.48\columnwidth]{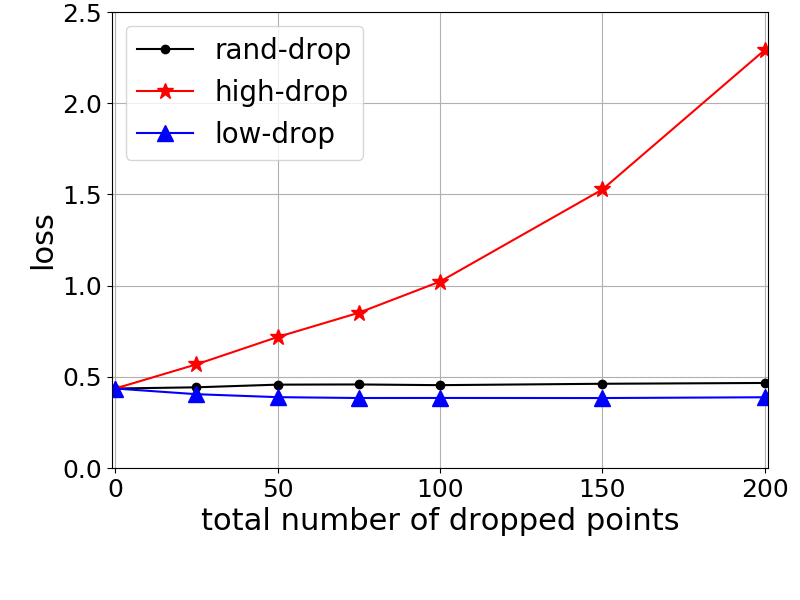}
	\includegraphics[width=0.48\columnwidth]{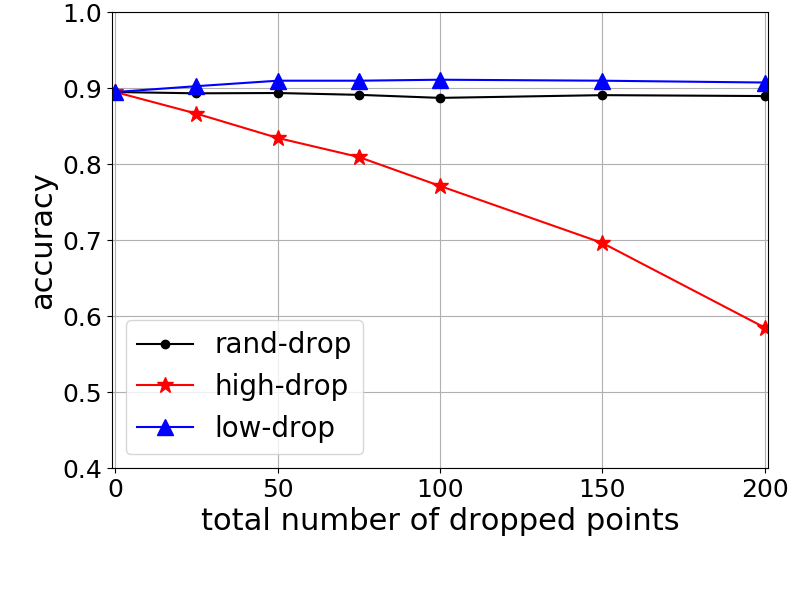}
	\vspace{-0.1cm}
	\caption{PointNet++ on 3D-MNIST and ModelNet40 from left to right: averaged loss (3D-MNIST), overall accuracy (3D-MNIST), averaged loss (ModelNet40), overall accuracy (ModelNet40).}
	\label{fig:pointnet2}
	\vspace{-0.2cm}
\end{figure*}
\begin{figure*}[t]	
	\centering
	\includegraphics[width=0.48\columnwidth]{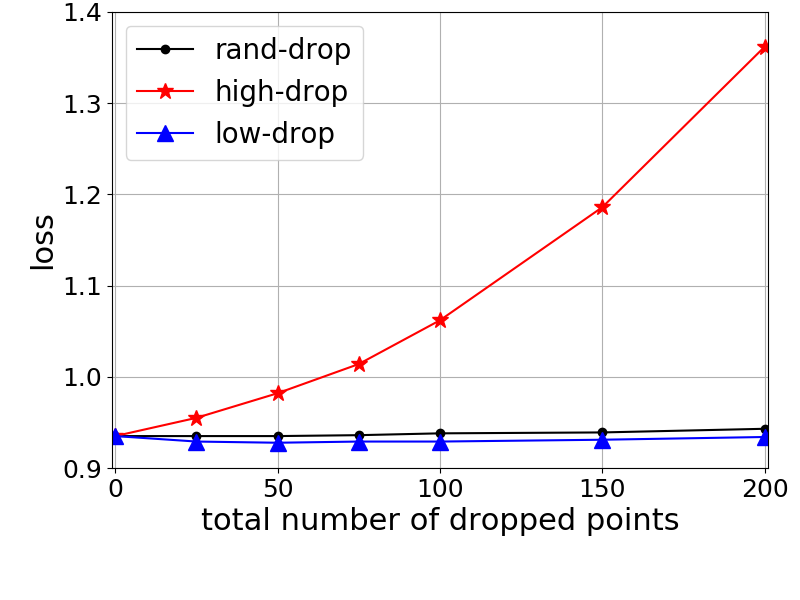}
	\includegraphics[width=0.48\columnwidth]{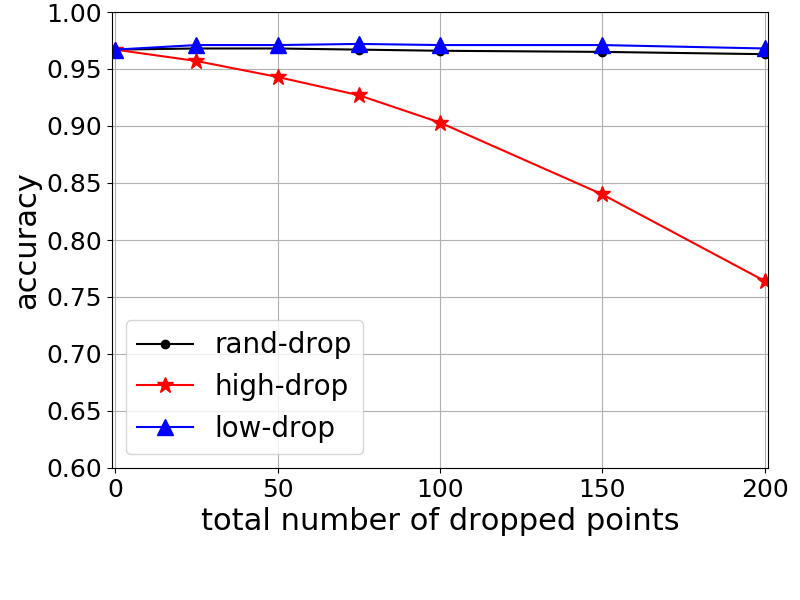}
	\includegraphics[width=0.48\columnwidth]{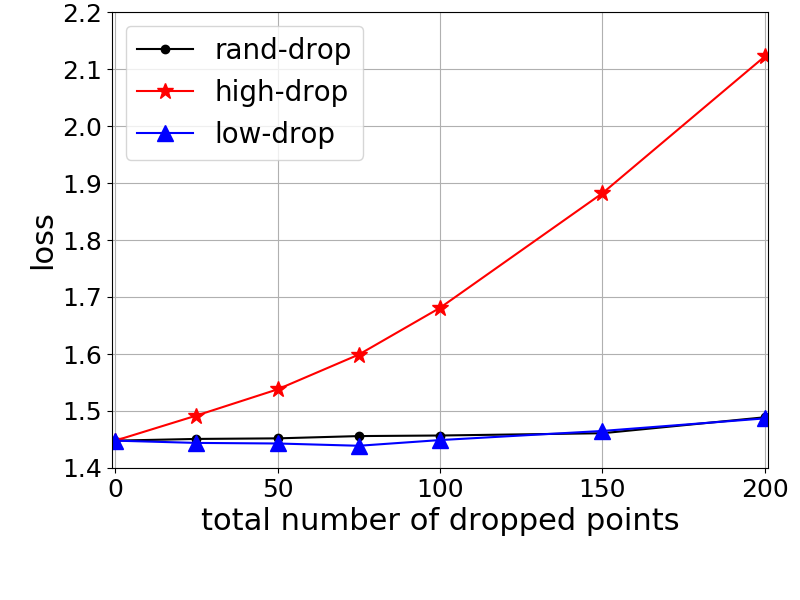}
	\includegraphics[width=0.48\columnwidth]{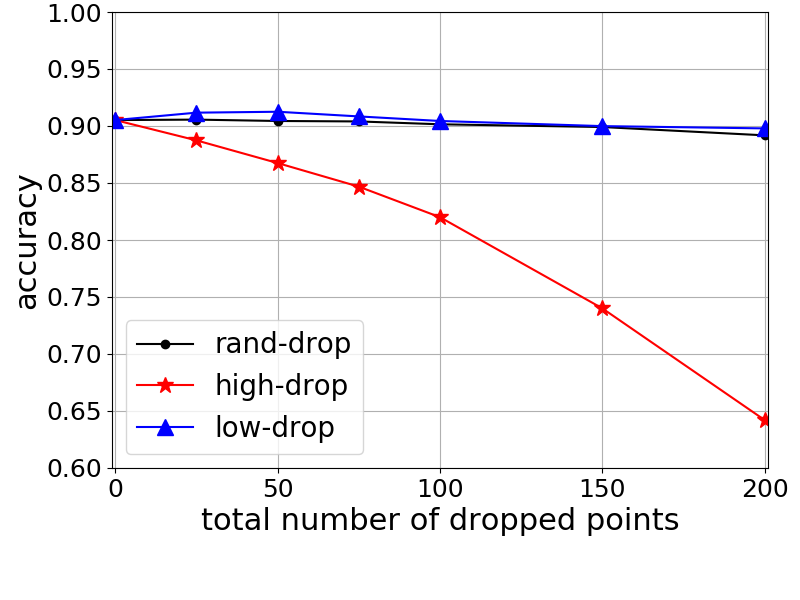}
	\vspace{-0.2cm}
	\caption{DGCNN on 3D-MNIST and ModelNet40: averaged loss (3D-MNIST), overall accuracy (3D-MNIST), averaged loss (ModelNet40), overall accuracy (ModelNet40).}
	\label{fig:dgcnn}
\end{figure*}

\begin{figure*}[t]	
	\centering
	\includegraphics[width=0.48\columnwidth]{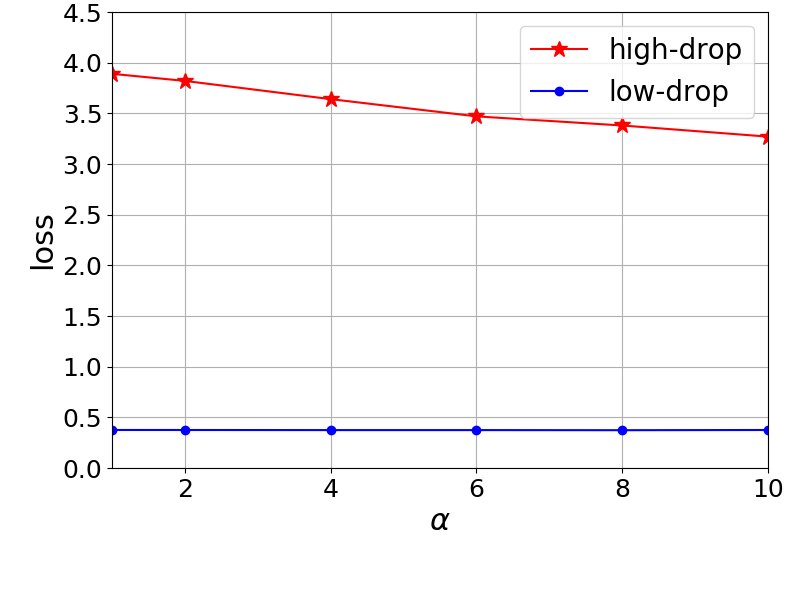}
	\includegraphics[width=0.48\columnwidth]{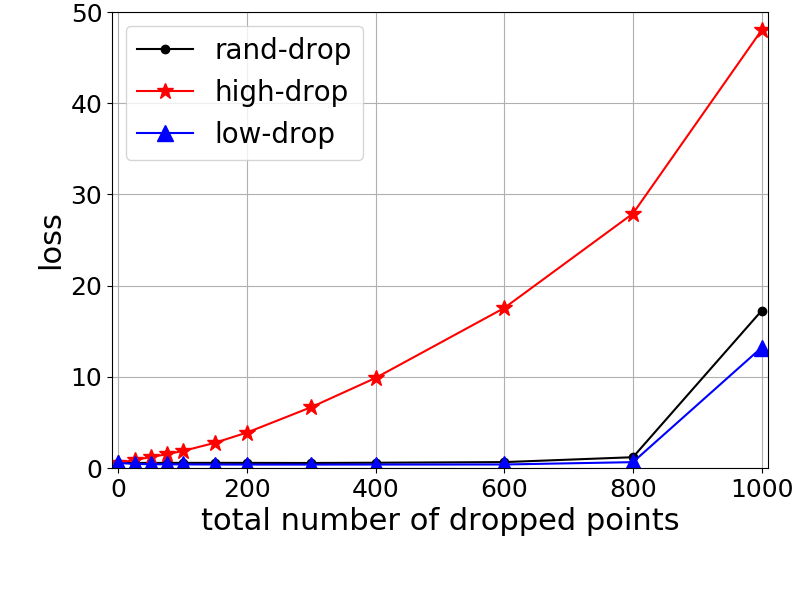}
	\includegraphics[width=0.48\columnwidth]{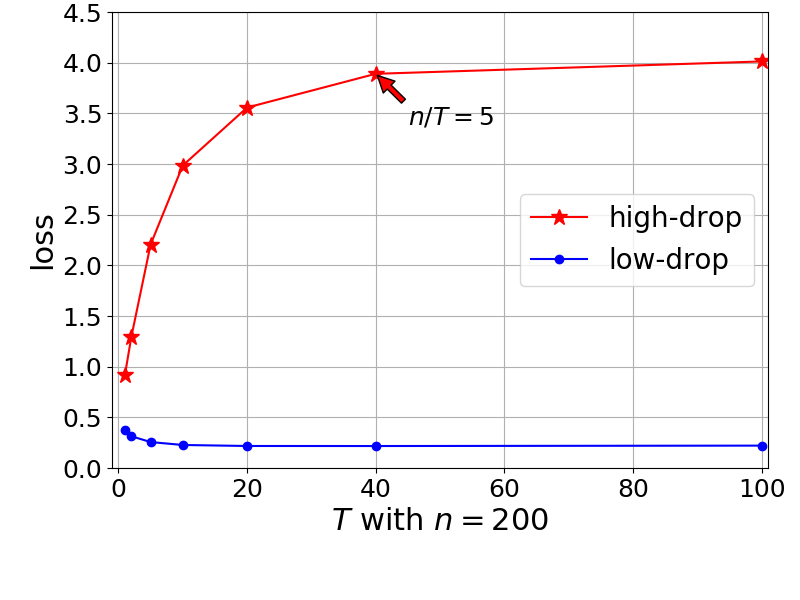}
	\includegraphics[width=0.48\columnwidth]{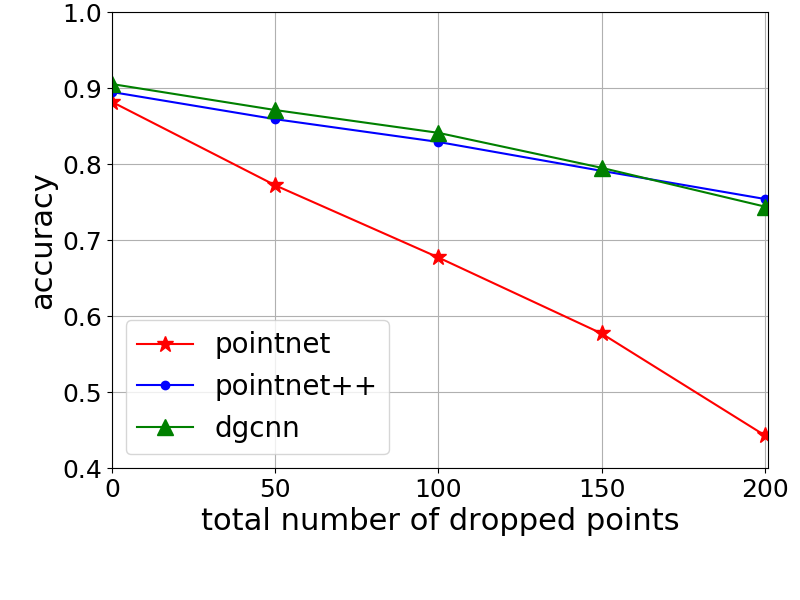}
	\caption{Impacts of hyper-parameters: (1) scaling factor $\alpha$, (2) number of dropped points $n$ (middle), (3) number of iterations $T$, (4) generalization results (subsets generated by point dropping on PointNet).}
	\label{fig:para}
	\vspace{-0.3cm}
\end{figure*}
\section{Experiments}
We verify our saliency map and point dropping algorithms by applying them to several benchmarks.
\subsection{Datasets and Models}
We use the two public datasets, 3D MNIST\footnote{\url{https://www.kaggle.com/daavoo/3d-mnist/version/13}} and ModelNet40\footnote{\url{http://modelnet.cs.princeton.edu/}} \cite{wu20153d}, to test our saliency map and point-dropping algorithms.
3D MNIST contains $6000$ raw 3D point clouds generated from 2D MNIST images, among which $5000$ are used for training and $1000$ for testing. Each raw point cloud contains about $20000$ 3D points. To enrich the dataset, we randomly select $1024$ points from each raw point cloud for 10 times to create 10 point clouds, making a training set of size $50000$ and a testing set of size $10000$, with each point cloud consisting of 1024 points. 
ModelNet40 contains 12,311 meshed CAD models of 40 categories, where 9,843 models are used for training and 2,468 models are for testing. We use the same point-cloud data provided by \cite{qi2017pointnet}, which are sampled from the surfaces of those CAD models. Finally, our approach is evaluated on state-of-the-art point cloud models introduced in section \ref{sec:3dmodel}, {\it i.e.}, PointNet, PointNet++ and DGCNN.
\begin{figure*}	
	\centering
	\includegraphics[width=1.9\columnwidth]{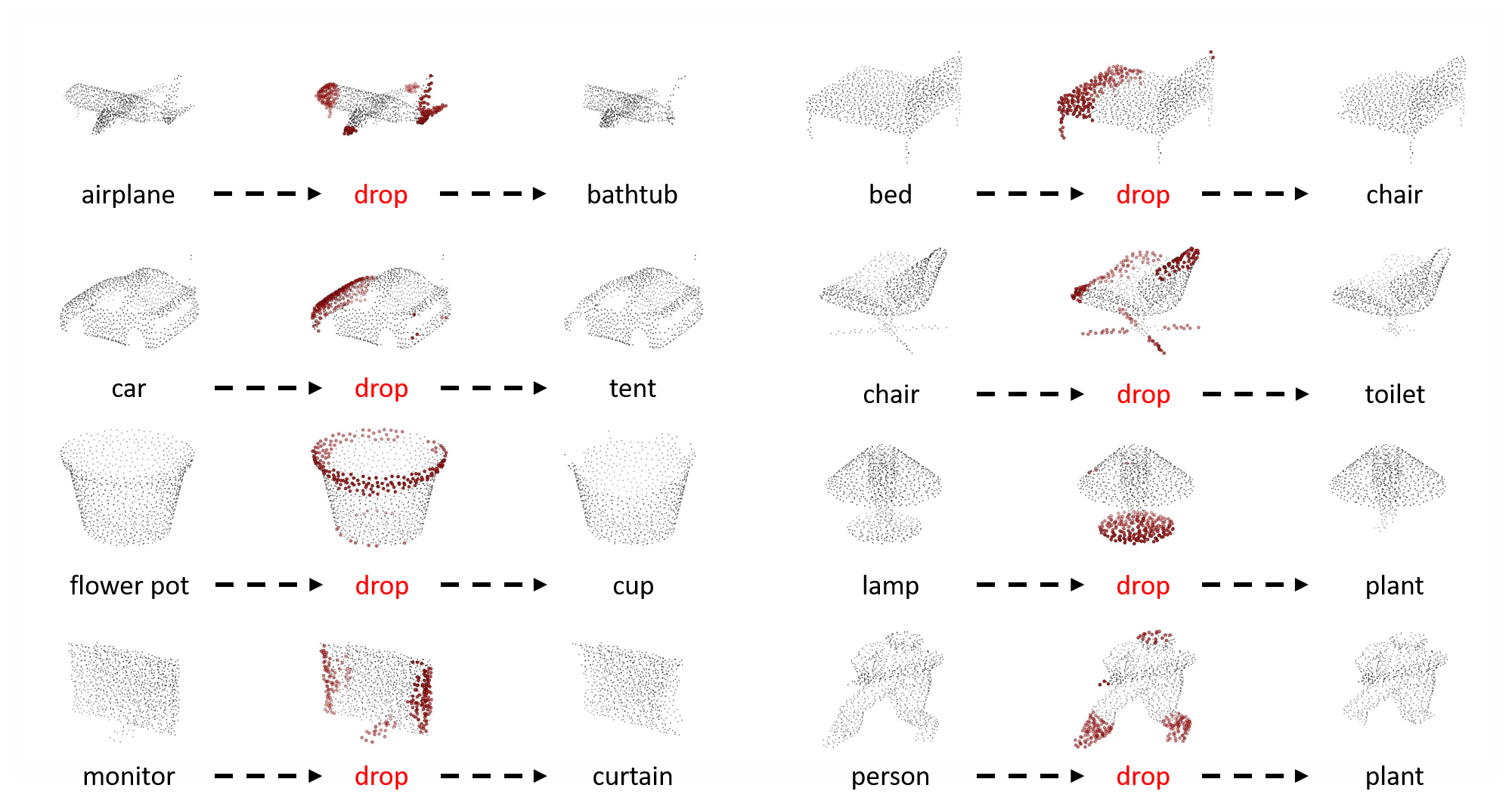}
	\caption{High-score point dropping (high-drop): original correct prediction (left), dropped points associated with {\em highest scores} by Algorithm~\ref{alg:iterdrop} (middle), wrong prediction after point dropping (right).}
	\label{fig:high}
	\vspace{-0.1cm}
\end{figure*}
\begin{figure*}	
	\centering
	\includegraphics[width=1.9\columnwidth]{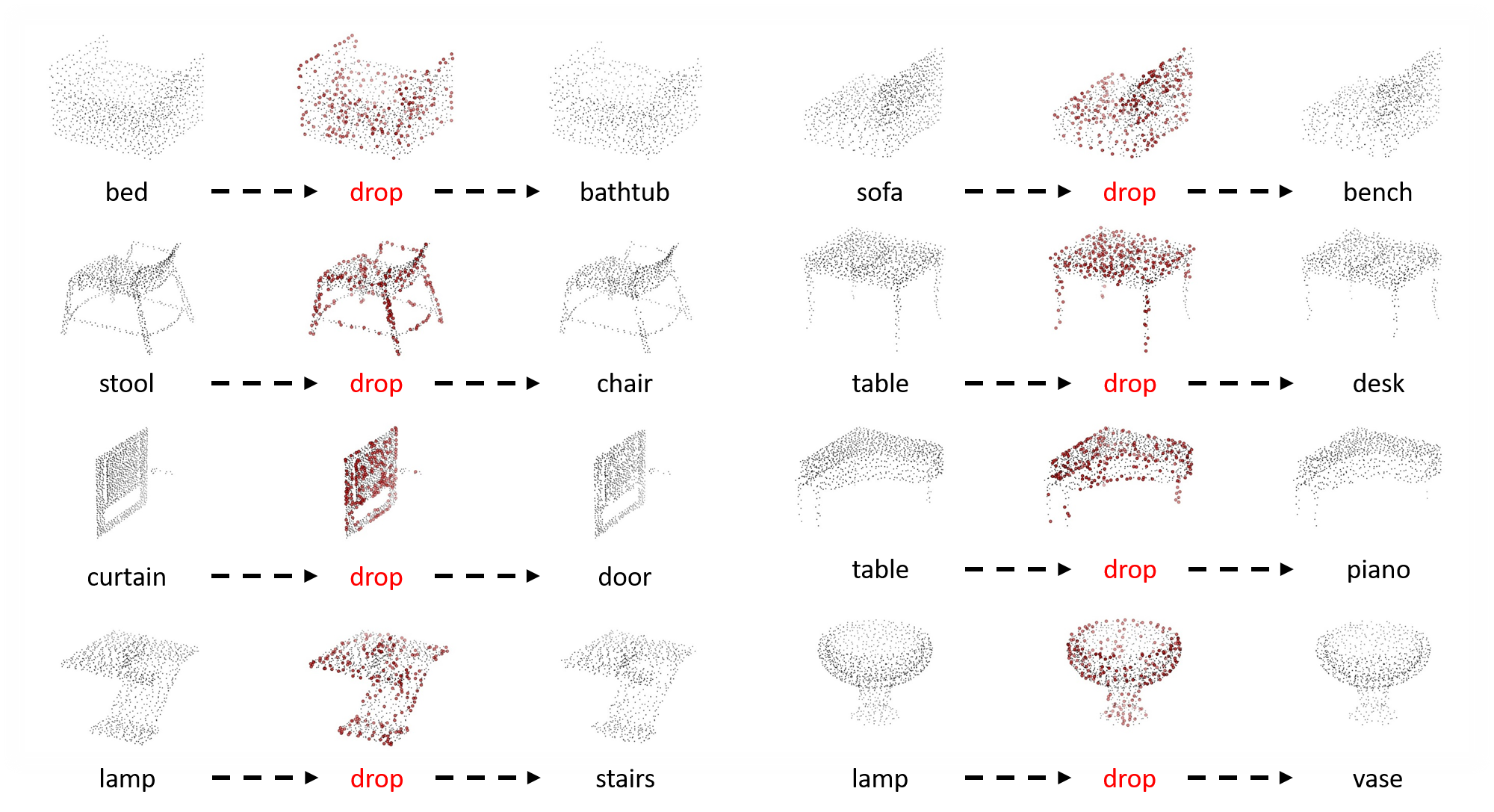}
	\caption{Low-score point dropping (low-drop): original wrong prediction (left), dropped points associated with {\em lowest} scores (middle), correct prediction after point dropping (right).}
	\label{fig:low}
	\vspace{-0.1cm}
\end{figure*}

\subsection{Implementation Details}
Our implementation is based on the models and code provided by \cite{qi2017pointnet,qi2017pointnet++,wang2018dynamic}
Default settings are used to train these models.
To enable dynamic point-number input along the second dimension of the batch-input tensor, for all the three models, we substitute several Tensorflow ops with equivalent ops that support dynamic inputs. We also rewrite a dynamic batch-gather ops and its gradient ops for DGCNN by C++ and Cuda. {\em For simplicity, we set the number of votes \footnote{Aggregate classification scores from multiple rotations} as 1. In all of the following cases, approximately $1\%$ accuracy improvement can be obtained by more votes, {\it e.g.}, 12 votes. Besides, incorporation of additional features like face normals will further improve the accuracy by nearly $1\%$.} We did not consider these tricks in our experiments for simplicity.

\subsection{Empirical Results}\label{sec:empirical}
To verify the veracity of our saliency map, we compare our saliency-map-driven point dropping approaches with the random point-dropping baseline \cite{qi2017pointnet}, denoted as {\em rand-drop}, and the critical-subset-based strategy introduced in Section \ref{subsec:subset_alg}, denoted as {\em critical} (only applicable to the PointNet structure). For simplicity, we refer to dropping $n$ points with the lowest saliency scores as {\em low-drop}, and dropping $n$ points with highest positive scores as {\em high-drop} in the followings. For {\em low-drop}, we found one iteration of Algorithm~\ref{alg:iterdrop} is already enough to achieve a good performance. While for {\em high-drop}, as explained in \ref{subsec:salient_alg}, we set $n/T = 5$ when dropping points with the highest scores in order to achieve better performance. We will further explain why we use this setting in the parameter study.
\vspace{-0.1cm}
\setcounter{footnote}{0}
\paragraph{Results on PointNet}
The performance of PointNet on 3D-MNIST test set is shown in Fig. \ref{fig:pointnet}. The overall accuracy of PointNet {\em maintains $94\% \sim 95\%$ } under rand-drop while varying the number of dropped points between 0 to 200. In contrast, {\em high-drop} reduces PointNet's overall accuracy to $49.2\%$. Furthermore, it is interesting to see by dropping points with negative scores, the accuracy even increases compared to using original point clouds by nearly $1\%$. This is consistent for other models and datasets as shown below.
For ModelNet40, as shown in Fig. \ref{fig:pointnet}, the overall accuracy of PointNet {\em maintains $87\% \sim 89\%$ }\footnote{$89.2\%$ in \cite{qi2017pointnet} can be acquired by setting the number of votes as $12$. We set the number of votes to $1$ for simplicity. The discrepancy between the accuracies under these two setting is always less than $1\%$.} under rand-drop. However, our point-dropping algorithm can increase/reduce the accuracy to $91.4\%$/$44.3\%$.
\vspace{-0.1cm}
\paragraph{Results on PointNet++}
The results for PointNet++ are shown in Fig. \ref{fig:pointnet2}, which maintains $95\%\sim96\%$ on 3D-MNIST under rand-drop, while our point-dropping algorithm can increase/reduce the accuracy to $97.2\%$/$59.5\%$. 
On the ModelNet40 test set, PointNet++ maintains $88 \sim 90\%$\footnote{$91.9\%$ in \cite{qi2017pointnet++} can be achieved by incorporating face normals as additional features and setting the number of votes as $12$} overall accuracy under rand-drop, while our algorithm can increase/reduce the accuracy to $91.1\%$/$58.5\%$.
\vspace{-0.1cm}
\paragraph{Results on DGCNN}
The accuracies of DGCNN on 3D-MNIST and ModelNet40 test sets are shown in Fig. \ref{fig:dgcnn}, respectively. Similarly, under rand-drop, DGCNN maintains $96\% \sim 97\%$ and $89\% \sim 91\%$ accuracies respectively. Given the same conditions, our algorithm is able to increase/reduce the accuracies to $97.2\%$/$76.4\%$ and $91.3\%$/$64.2\%$ respectively.
\vspace{-0.1cm}
\paragraph{Visualization}
Several point clouds manipulated by {\em high-drop} are visualized in Fig.~\ref{fig:high}. For the point clouds shown in those figures, our saliency map and the iterative point-dropping algorithm successfully identify the important segments (i.e., the dropped segments) that distinguish them from other clouds, {\it e.g.,} the base of the lamp. It is worth pointing out that human also recognize several point clouds in Fig.~\ref{fig:high} as other objects. On the contrary, as shown in Fig.~\ref{fig:low}, {\em low-drop} is visually similar to a denoising process, {\it i.e.}, dropping noisy/useless points scattered throughout point clouds. Although the DNN model misclassifies the original point clouds in some cases, dropping those noisy points could correct the model predictions.
\vspace{-0.1cm}
\paragraph{Parameter Study}
We employ PointNet on ModelNet40 to study the impacts of the scaling factor $\alpha$, the number of dropped points $n$, and the number of iterations $T$ to model performance. As shown in Fig.~\ref{fig:para}, $\alpha = 1$ is a good setting for Algorithm \ref{alg:iterdrop} since as $\alpha$ increases, model prediction loss will slightly decrease. Besides, it is clear in Fig.~\ref{fig:para} (2nd) that our {\em high-drop} significantly outperforms {\em rand-drop} in terms of degrading model performance: the accuracy of PointNet still maintains over $80\%$ under rand-drop with $600/1024$ points dropped, while {\em high-drop} reduces the accuracy to nearly $0$. In Fig.~\ref{fig:para} (3rd), we show that more iterations lead to better performance. However, when it comes to {\em low-drop}, more iterations only slightly improve the performance but with more computational cost. Therefore, we recommend executing our algorithm for 20 iterations to identify the important subsets (high-drop), and for one iteration to denoise the point clouds (low-drop).
\vspace{-0.1cm}
\paragraph{Generalization}
{\em We also show the generalization performance of our algorithm in Fig.~\ref{fig:para} (4th). Specifically, we test the PointNet-generated subsets (after dropping high-score
points) on the PointNet++ and DGCNN, and the accuracy still degrades a lot.}
\vspace{-0.1cm}
\paragraph{Discussion}
Among all the three state-of-the-art DNN models for 3D point clouds, DGCNN appears to be the most robust model to point dropping (missing), which indicates DGCNN depends more on the entire point cloud rather than certain point or segment. We conjecture the robustness comes from its structures designed to capture more local and global information, which is supposed to compensate for the information loss by dropping points or segments. On the contrary, PointNet does not capture local structures \cite{qi2017pointnet++}, making it the most sensitive model to point dropping.

\section{Conclusion}
In this paper, a saliency-map is constructed for 3D point-clouds to measure the contribution (importance) of each point in a point cloud to model prediction loss. By approximating point dropping with a continuous point-shifting procedure, we show that the contribution of a point is approximately proportional to, and thus can be scored by, the gradient of loss w.r.t.\! the point under a scaled spherical-coordinate system. Using this saliency map, we further standardize the point-dropping process to verify the veracity of our saliency map on characterizing point-level and subset-level saliency.
Extensive evaluations show that our saliency-map-driven point-dropping algorithm consistently outperforms other schemes such as the random point-dropping scheme and critical-subset based strategy, indicating that our saliency is a more accurate measure to quantify the point-level and subset-level saliency of a point cloud.
{\small
\bibliographystyle{ieee_fullname}
\bibliography{egpaper_final}

\begin{thebibliography}{10}\itemsep=-1pt

\bibitem{biswas2012depth}
Joydeep Biswas and Manuela Veloso.
\newblock Depth camera based indoor mobile robot localization and navigation.
\newblock In {\em 2012 IEEE International Conference on Robotics and
  Automation}, pages 1697--1702. IEEE, 2012.

\bibitem{bohm2008outlier}
Christian B{\"o}hm, Christos Faloutsos, and Claudia Plant.
\newblock Outlier-robust clustering using independent components.
\newblock In {\em Proceedings of the 2008 ACM SIGMOD international conference
  on Management of data}, pages 185--198. ACM, 2008.

\bibitem{hadsell2009learning}
Raia Hadsell, Pierre Sermanet, Jan Ben, Ayse Erkan, Marco Scoffier, Koray
  Kavukcuoglu, Urs Muller, and Yann LeCun.
\newblock Learning long-range vision for autonomous off-road driving.
\newblock {\em Journal of Field Robotics}, 26(2):120--144, 2009.

\bibitem{kanezaki2018rotationnet}
Asako Kanezaki, Yasuyuki Matsushita, and Yoshifumi Nishida.
\newblock Rotationnet: Joint object categorization and pose estimation using
  multiviews from unsupervised viewpoints.
\newblock In {\em Proceedings of IEEE International Conference on Computer
  Vision and Pattern Recognition (CVPR)}, 2018.

\bibitem{kehoe2013cloud}
Ben Kehoe, Akihiro Matsukawa, Sal Candido, James Kuffner, and Ken Goldberg.
\newblock Cloud-based robot grasping with the google object recognition engine.
\newblock In {\em 2013 IEEE International Conference on Robotics and
  Automation}, pages 4263--4270. IEEE, 2013.

\bibitem{linsen2001point}
Lars Linsen.
\newblock {\em Point cloud representation}.
\newblock Univ., Fak. f{\"u}r Informatik, Bibliothek Technical Report, Faculty
  of Computer~…, 2001.

\bibitem{maturana2015voxnet}
Daniel Maturana and Sebastian Scherer.
\newblock Voxnet: A 3d convolutional neural network for real-time object
  recognition.
\newblock In {\em Intelligent Robots and Systems (IROS), 2015 IEEE/RSJ
  International Conference on}, pages 922--928. IEEE, 2015.

\bibitem{papernot2016limitations}
Nicolas Papernot, Patrick McDaniel, Somesh Jha, Matt Fredrikson, Z~Berkay
  Celik, and Ananthram Swami.
\newblock The limitations of deep learning in adversarial settings.
\newblock In {\em Security and Privacy (EuroS\&P), 2016 IEEE European Symposium
  on}, pages 372--387. IEEE, 2016.

\bibitem{qi2017pointnet}
Charles~R Qi, Hao Su, Kaichun Mo, and Leonidas~J Guibas.
\newblock Pointnet: Deep learning on point sets for 3d classification and
  segmentation.
\newblock {\em Proc. Computer Vision and Pattern Recognition (CVPR), IEEE},
  1(2):4, 2017.

\bibitem{qi2017pointnet++}
Charles~Ruizhongtai Qi, Li Yi, Hao Su, and Leonidas~J Guibas.
\newblock Pointnet++: Deep hierarchical feature learning on point sets in a
  metric space.
\newblock In {\em Advances in Neural Information Processing Systems}, pages
  5099--5108, 2017.

\bibitem{rusu2009fast}
Radu~Bogdan Rusu, Nico Blodow, and Michael Beetz.
\newblock Fast point feature histograms (fpfh) for 3d registration.
\newblock In {\em 2009 IEEE International Conference on Robotics and
  Automation}, pages 3212--3217. IEEE, 2009.

\bibitem{rusu2008towards}
Radu~Bogdan Rusu, Zoltan~Csaba Marton, Nico Blodow, Mihai Dolha, and Michael
  Beetz.
\newblock Towards 3d point cloud based object maps for household environments.
\newblock {\em Robotics and Autonomous Systems}, 56(11):927--941, 2008.

\bibitem{simonyan2013deep}
Karen Simonyan, Andrea Vedaldi, and Andrew Zisserman.
\newblock Deep inside convolutional networks: Visualising image classification
  models and saliency maps.
\newblock {\em arXiv preprint arXiv:1312.6034}, 2013.

\bibitem{springenberg2014striving}
Jost~Tobias Springenberg, Alexey Dosovitskiy, Thomas Brox, and Martin
  Riedmiller.
\newblock Striving for simplicity: The all convolutional net.
\newblock {\em arXiv preprint arXiv:1412.6806}, 2014.

\bibitem{su2015multi}
Hang Su, Subhransu Maji, Evangelos Kalogerakis, and Erik Learned-Miller.
\newblock Multi-view convolutional neural networks for 3d shape recognition.
\newblock In {\em Proceedings of the IEEE international conference on computer
  vision}, pages 945--953, 2015.

\bibitem{thrun2006probabilistic}
Sebastian Thrun, Michael Montemerlo, and Andrei Aron.
\newblock Probabilistic terrain analysis for high-speed desert driving.
\newblock In {\em Robotics: Science and Systems}, pages 16--19, 2006.

\bibitem{vosselman2004recognising}
George Vosselman, Ben~GH Gorte, George Sithole, and Tahir Rabbani.
\newblock Recognising structure in laser scanner point clouds.
\newblock {\em International archives of photogrammetry, remote sensing and
  spatial information sciences}, 46(8):33--38, 2004.

\bibitem{wang2017dominant}
Chu Wang, Marcello Pelillo, and Kaleem Siddiqi.
\newblock Dominant set clustering and pooling for multi-view 3d object
  recognition.
\newblock In {\em Proceedings of British Machine Vision Conference (BMVC)},
  volume~12, 2017.

\bibitem{wang2018dynamic}
Yue Wang, Yongbin Sun, Ziwei Liu, Sanjay~E Sarma, Michael~M Bronstein, and
  Justin~M Solomon.
\newblock Dynamic graph cnn for learning on point clouds.
\newblock {\em arXiv preprint arXiv:1801.07829}, 2018.

\bibitem{wu20153d}
Zhirong Wu, Shuran Song, Aditya Khosla, Fisher Yu, Linguang Zhang, Xiaoou Tang,
  and Jianxiong Xiao.
\newblock 3d shapenets: A deep representation for volumetric shapes.
\newblock In {\em Proceedings of the IEEE conference on computer vision and
  pattern recognition}, pages 1912--1920, 2015.

\bibitem{yu2018multi}
Tan Yu, Jingjing Meng, and Junsong Yuan.
\newblock Multi-view harmonized bilinear network for 3d object recognition.
\newblock In {\em Proceedings of the IEEE Conference on Computer Vision and
  Pattern Recognition}, pages 186--194, 2018.

\end{thebibliography}
}

\end{document}